\documentclass[11pt,a4paper]{article}
\usepackage[hyperref]{emnlp2020}
\usepackage{times}
\usepackage{latexsym}

\usepackage{microtype}

\aclfinalcopy 

\usepackage{amsmath}
\usepackage{amssymb}

\usepackage{url}
\usepackage{comment}
\newcommand{\parheader}[1]{{\smallskip \noindent \bf #1}}
\newcommand{\parheaderwithdot}[1]{{\smallskip \noindent \bf #1.}}
\usepackage{algorithm}
\usepackage{algpseudocode}
\usepackage{graphicx}
\usepackage{multirow}
\usepackage{booktabs}

\title{MKD: a Multi-Task Knowledge Distillation Approach \\ for Pretrained Language Models}

\author{Linqing Liu,\thanks{All work was done while the first author was a research intern at Salesforce Research.} $^1$ Huan Wang,$^2$ Jimmy Lin,$^1$  Richard Socher,$^2$ \and Caiming Xiong$^2$ \vspace{0.1cm}\\
$^1$ David R. Cheriton School of Computer Science, University of Waterloo \\ $^2$ Salesforce Research \\
{\small \tt  \{linqing.liu,\hspace{.15em}jimmylin\}@uwaterloo.ca, \{huan.wang,\hspace{.15em}rsocher,\hspace{.15em}cxiong\}@salesforce.com}
}

\date{}

\begin{document}
\maketitle
\begin{abstract}
Pretrained language models have led to significant performance gains in many NLP tasks. However, the intensive computing resources to train such models remain an issue. Knowledge distillation alleviates this problem by learning a light-weight student model. So far the distillation approaches are all task-specific. In this paper, we explore knowledge distillation under the multi-task learning setting. The student is jointly distilled across different tasks. It acquires more general representation capacity through multi-tasking distillation and can be further fine-tuned to improve the model in the target domain. Unlike other BERT distillation methods which specifically designed for Transformer-based architectures, we provide a general learning framework. Our approach is model agnostic and can be easily applied on different future teacher model architectures. We evaluate our approach on a Transformer-based and LSTM based student model. Compared to a strong, similarly LSTM-based approach, we achieve better quality under the same computational constraints. Compared to the present state of the art, we reach comparable results with much faster inference speed.
\end{abstract}

\section{Introduction}
Pretrained language models learn highly effective language representations from large-scale unlabeled data. A few prominent examples include ELMo~\cite{peters2018deep}, BERT~\cite{devlin2019bert}, RoBERTa~\cite{liu2019roberta}, and XLNet~\cite{yang2019xlnet}, all of which have achieved state of the art in many natural language processing (NLP) tasks, such as natural language inference, sentiment classification, and semantic textual similarity. However, such models use dozens, if not hundreds, of millions of parameters, invariably leading to resource-intensive inference. The consensus is that we need to cut down the model size and reduce the computational cost while maintaining comparable quality.

One approach to address this problem is knowledge distillation (KD; \citealp{ba2014deep, hinton2015distilling}), where a large model functions as a \textit{teacher} and transfers its knowledge to a small \textit{student} model. Previous methods focus on task-specific KD, which transfers knowledge from a single-task teacher to its single-task student. Put it another way, the knowledge distillation process needs to be conducted all over again when  performing on a new NLP task. The inference speed of the large-scale teacher model remains the bottleneck for various downstream tasks distillation.

\begin{figure}
    \centering
    \includegraphics[scale=0.4]{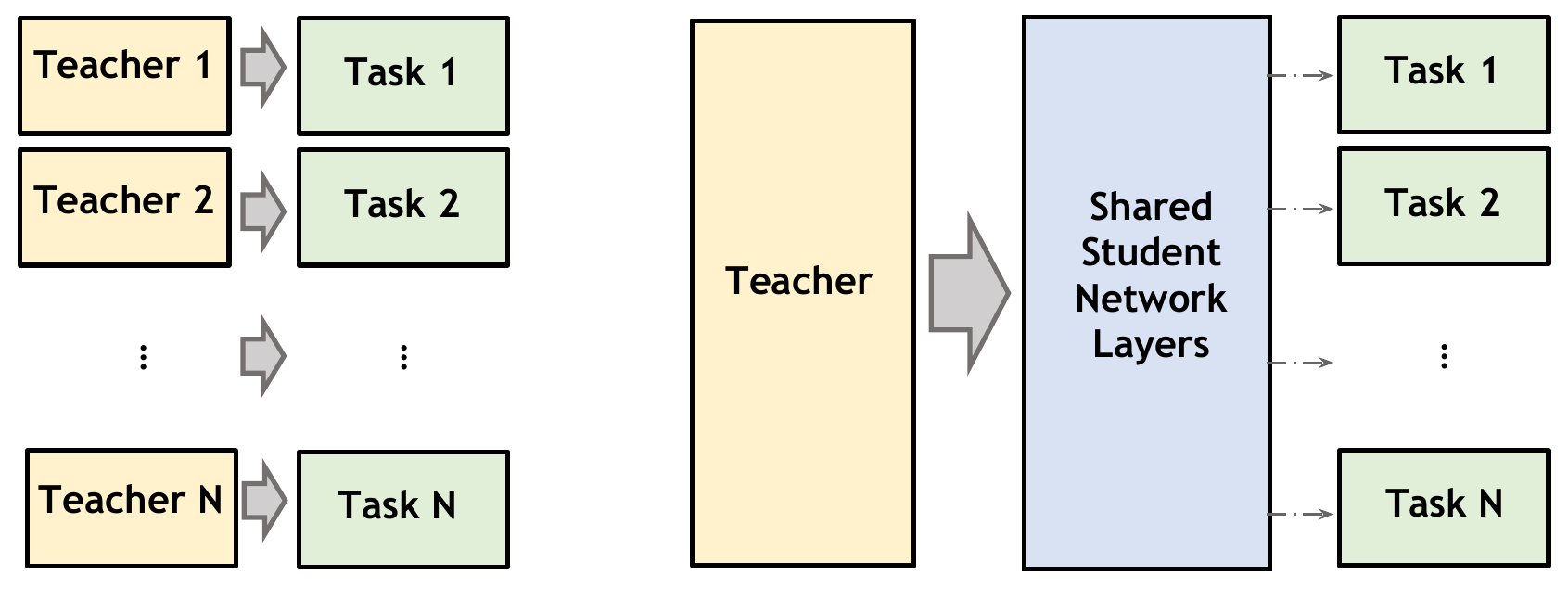}
    \caption{The left figure represents task-specific KD. The distillation process needs to be performed for each different task. The right figure represents our proposed multi-task KD. The student model consists of shared layers and task-specific layers.}
    \label{fig:mtlkd_archt}
\end{figure}

Our goal is to find a \textit{distill-once-fits-many} solution. In this paper, we explore the knowledge distillation method under the setting of multi-task learning~(MTL; \citealp{caruana1997multitask, baxter2000model}). We propose to distill the student model from different tasks jointly. The overall framework is illustrated in Figure \ref{fig:mtlkd_archt}. The reason is twofold: first, the distilled model learns a more universal language representation by leveraging cross-task data. Second, the student model achieves both comparable quality and fast inference speed across multiple tasks. MTL is based on the idea \cite{maurer2016benefit} that tasks are related by means of a common low dimensional representation. We also provide an intuitive explanation on why using shared structure could possibly help by assuming some connections over the conditional distribution of different tasks.

We evaluate our approach on two different student model architectures. One uses three layers Transformers \cite{vaswani2017attention}, since most of the KD works \cite{sun2019patient, jiao2019tinybert} use Transformers as their students. Another is LSTM based network with bi-attention mechanism. Previously \citet{tang2019distilling} examine the representation capacity of a simple, single-layer Bi-LSTM only, so we are interested in whether adding more previous effective modules, such as an attention mechanism, will further improve its effectiveness. 
It exemplifies that our approach is model agnostic, i.e., the choice of student model does not depend on the teacher model architecture; The teacher model can be easily switched to other powerful language models other than BERT.

We further study several important problems in knowledge distillation, such as the choice of modules in student model, the influence of different tokenization methods, and the influence of MTL in KD. We evaluate our approach on seven datasets across four different tasks. For LSTM based student, our approach keeps the advantage of inference speed while maintaining comparable performances as those specifically designed for Transformer methods. For our Transformer based student, it does provide a modest gain, and outperforms other KD methods without using external training data.

\section{Related Work}
\parheaderwithdot{Language model pretraining}
Given a sequence of tokens, pretrained language models encode each token as a general language representational embedding. A large body of literature has explored this area. Traditional pretrained word representations~\cite{turian2010word} presume singular word meanings and thus adapt poorly to multiple contexts---for some notable examples, see word2vec \cite{mikolov2013efficient}, GloVe \cite{pennington2014glove}, and FastText \cite{bojanowski2017enriching}. For more flexible word representations, a few advancements exist:~\citet{neelakantan2015efficient} learn multiple embeddings per word type;~context2vec~\cite{melamud2016context2vec} uses bidirectional LSTM to encode contexts around target words;~CoVe \cite{mccann2017learned} trains LSTM encoders on some machine translation datasets, showing that these encoders are well-transferable to other tasks. Prominently, ELMo \cite{peters2018deep} learns deep word representations using a bidirectional language model. It can be easily added to an existing model and boost performance across six challenging NLP tasks.

Fine-tuning approaches are mostly employed in more recent work. They pretrain the language model on a large-scale unlabeled corpus and then fine-tune it with in-domain labeled data for a supervised downstream task~\cite{dai2015semi, howard2018universal}. BERT \cite{devlin2019bert}, GPT \cite{radford2018improving} and GPT-2~\cite{radford2019language} are some of the prominent examples. Following BERT, XLNet \cite{yang2019xlnet} proposes a generalized autoregressive pretraining method and RoBERTa \cite{liu2019roberta}  optimizes BERT pretraining approach. These pretrained models are large in size and contain millions of parameters. We target the BERT model and aim to address this problem through knowledge distillation. Our approach can be easily applied to other models as well.

\parheaderwithdot{Knowledge distillation}
Knowledge distillation \cite{ba2014deep, hinton2015distilling} transfers knowledge from a large \textit{teacher} model to a smaller \textit{student} model. Since the distillation only matches the output distribution, the student model architecture can be completely different from that of the teacher model. There are already many efforts trying to distill BERT into smaller models. BERT-PKD \cite{sun2019patient} extracts knowledge not only from the last layer of the teacher, but also from previous layers. TinyBERT \cite{jiao2019tinybert} introduces a two-stage learning framework which performs transformer distillation at both pretraining and task-specific stages. \citet{zhao2019extreme} train a student model with smaller vocabulary and lower hidden states dimensions. DistilBERT \cite{sanh2019distilbert} reduces the layers of BERT and uses this small version of BERT as its student model. All the aforementioned distillation methods are performed on a single task, specifically designed for the transformer-based teacher architecture, resulting in poor generalizability to other type of models.
Our objective is to invent a general distillation framework, applicable to either transformer-based models or other architectures as well. \citet{tang2019distilling} distill BERT into a single-layer BiLSTM. In our paper, we hope to extract more knowledge from BERT through multi-task learning, while keeping the student model simple.

\parheaderwithdot{Multi-task learning} Multi-task learning (MTL) has been successfully applied on different applications \cite{collobert2008unified, deng2013new, girshick2015fast}. MTL helps the pretrained language models learn more generalized text representation by sharing the domain-specific information contained in each related task training signal~\cite{caruana1997multitask}. ~\citet{liu2019multi, liu2015representation} propose a multi-task deep neural network (MT-DNN) for learning representations across multiple tasks. \cite{clark2019bam} propose to use knowledge distillation so that single task models can teach a multi-task model. \citet{liu2019improving} train an ensemble of large DNNs and then distill their knowledge to a single DNN via multi-task learning to ensemble its teacher performance.

\section{Model Architecture}
In this section, we introduce the teacher model and student model for our distillation approach. We explored two different student architectures: a traditional bidirectional long short-term memory network (BiLSTM) with bi-attention mechanism in \ref{sec:3.2}, and the popular Transformer in \ref{sec:3.3}.

\subsection{Multi-Task Refined Teacher Model}\label{sec:3.1}
We argue that multi-task learning can leverage the regularization of different natural language understanding tasks. Under this setting, language models can be more effective in learning universal language representations. To this end, we consider the bidirectional transformer language model (BERT; \citealp{devlin2019bert}) as bottom shared text encoding layers, and fine-tune the task-specific top layers for each type of NLU task. There are mainly two stages for the training procedure: pretraining the shared layer and multi-task refining.

\parheaderwithdot{Shared layer pretraining}
Following \citet{devlin2019bert}, the input token is first encoded as the the summation of its corresponding token embeddings, segmentation embeddings and position embeddings. The input embeddings are then mapped into contextual embeddings $C$ through a multi-layer bidirectional transformer encoder. The pretraining of these shared layers use the cloze task and next sentence prediction task. We use the pretrained $BERT_{LARGE}$ to initialize these shared layers.

\parheaderwithdot{Multi-task refining}
The contextual embeddings ${C}$ are then passed through the upper task-specific layers. Following \citet{liu2019multi}, our current NLU training tasks on GLUE \cite{wang2018glue} can be classified into four categories: single-sentence classification (CoLA and SST-2), pairwise text classification (RTE, MNLI, WNLI, QQP, and MRPC), pairwise text similarity (STS-B), and relevance ranking (QNLI). Each category corresponds to its own output layer. 

Here we take the text similarity task as an example to demonstrate the implementation details. Following \citet{devlin2019bert}, we consider the contextual embedding of the special \texttt{[CLS]} token as the semantic representation of the input sentence pair ${(X_{1}, X_{2})}$. The similarity score can be predicted by the similarity ranking layer: 
\begin{equation}
    \text{Sim}(X_{1}, X_{2}) = W_{STS}^\top x
\end{equation}
where $W_{STS}$ is a task-specific learnable weight vector and $x$ is the contextual embedding of the \texttt{[CLS]} token.

In the multi-task refining stage, all the model parameters, including bottom shared layers and task-specific layers, are updated through mini-batch stochastic gradient descent \cite{li2014efficient}. The training data are packed into mini-batches and each mini-batch only contains samples from one task. Running all the mini-batches in each epoch approximately optimizes the sum all of all multi-task objectives. In each epoch, the model is updated according to the selected mini-batch and its task-specific objective. We still take the text similarity task as an example, where each pair of sentences is labeled with a real-value similarity score $y$. We use the mean-squared error loss as our objective function:
\begin{equation}
     \lVert y-\text{Sim}(X_{1}, X_{2}) \rVert^{2}_{2}
\end{equation}
For text classification task, we use the cross-entropy loss as the objective function. For relevance ranking task, we minimize the negative log likelihood of the positive examples \cite{liu2019multi}. We can also easily add other tasks by adding its own task-specific layer.

\subsection{LSTM-based Student Model}\label{sec:3.2}
\begin{figure}
    \centering
    \includegraphics[scale=0.4]{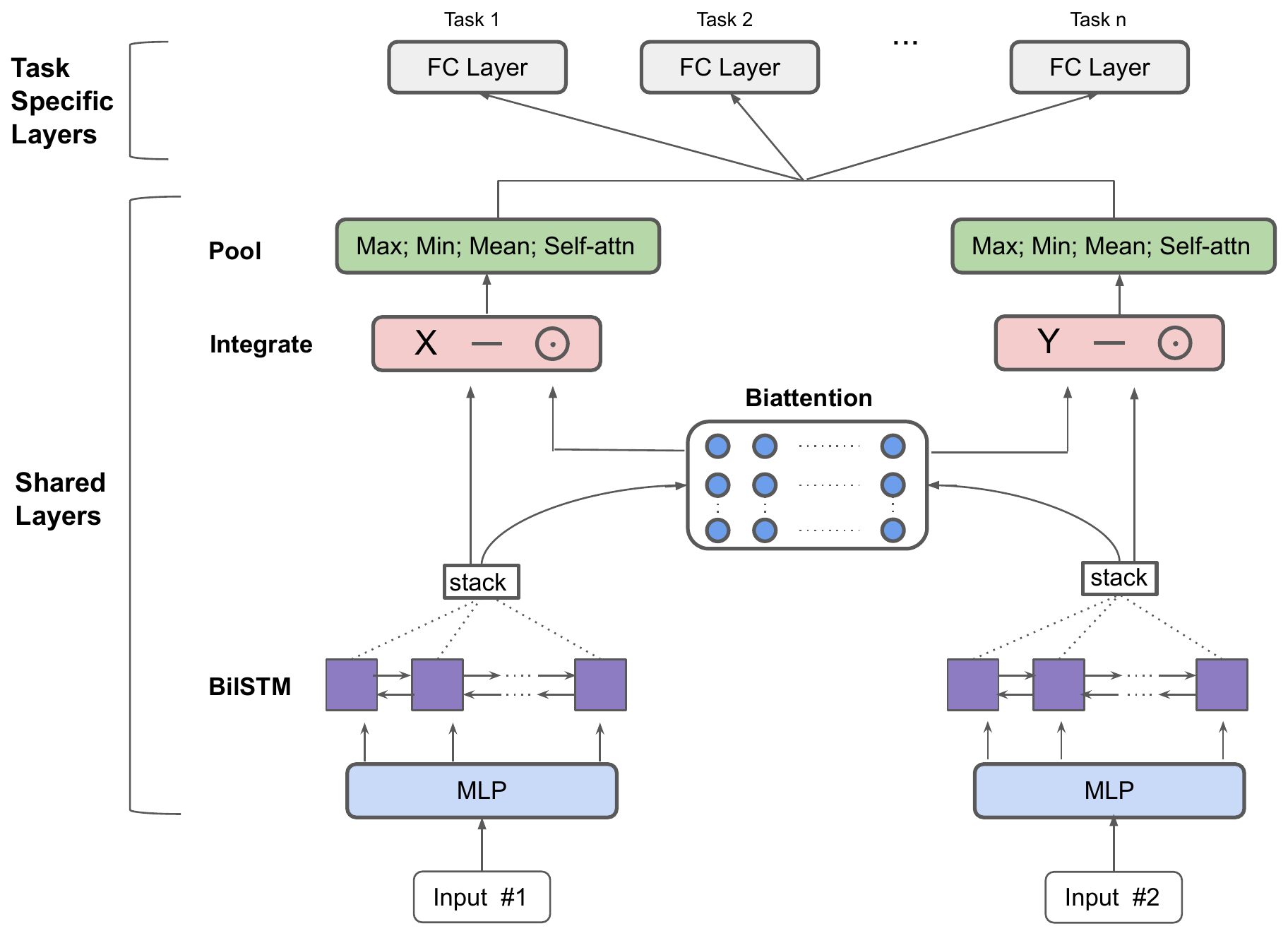}
    \caption{Architecture for the bi-attentive student neural network.}
    \label{fig:bcn}
\end{figure}
We're interested in exploring whether a simple architecture, such as LSTM, has enough representation capacity to transfer knowledge from the teacher model. We also incorporate bi-attention module since it's widely used between pairs of sentences \cite{peters2018deep, wang2018glue}. And the inputs in our experiments are mostly two sentences. Our LSTM-based bi-attentive student model is depicted in Figure \ref{fig:bcn}. For equation representations, the embedding vectors of input sequences are denoted as ${w^{x}}$ and ${w^{y}}$. For single-sentence input tasks, ${w^{y}}$ is the same as ${w^{x}}$. $\oplus$ represents vectors concatenation.

${w^{x}}$ and ${w^{y}}$ are first converted into ${\hat{w}^{x}}$ and ${\hat{w}^{y}}$ through a feedforward network with ReLU activation \cite{nair2010rectified} function. For each token in ${\hat{w}^{x}}$ and ${\hat{w}^{y}}$, we then use a bi-directional LSTM encoder to compute its hidden states and stack them over time axis to form matrices $X$ and $Y$ separately.

Next, we apply the biattention mechanism \cite{xiong2016dynamic, seo2016bidirectional} to compute the attention contexts $A=XY^\top$ of the input sequences. The attention weight $A_{x}$ and $A_{y}$ is extracted through a column-wise normalization for each sequence. The context vectors $C_{x}$ and $C_{y}$  for each token is computed as the multiplication of its corresponding representation and attention weight:
\begin{equation}
    A_{x} = \text{softmax}(A) \quad C_{x} = A_{x}^\top X
\end{equation}

Same as \cite{mccann2017learned}, we concatenate three different computations between original representations and context vector to reinforce their relationships. The concatenated vectors are then passed through one single-layer BiLSTM:
\begin{equation}
    \begin{split}
        & X_{y} = \text{BiLSTM}([X \oplus X - C_{y} \oplus X \odot C_{y}]) \\
        & Y_{x} = \text{BiLSTM}([Y \oplus Y - C_{x} \oplus Y \odot C_{x}])
    \end{split}
\end{equation}
The pooling operations are then applied on the outputs of BiLSTM. We use max, mean, and self-attentive pooling to extract features. These three pooled representations are then concatenated to get one context representation. We feed this context representation through a fully-connected layer to get final output.

\subsection{Transformer-based Student Model}\label{sec:3.3}
Most of the pre-trained language models, which can be employed as teachers, are built with Transformers. Transformer \cite{vaswani2017attention} now is an ubiquitous model architecture. It draws global dependencies between input and output entirely relying on an self-attention mechanism. Our student model uses three layers of Transformers. Same as BERT \cite{devlin2019bert}, [CLS] is added in front of every input example, and [SEP] is added between two input sentences. We use the average [CLS] representation from each layer as the final output.

\section{Multi-Task Distillation}
The parameters of student models, introduced in Section \ref{sec:3.2} and \ref{sec:3.3}, are shared across all tasks. Each task has its individual layer on top of it. We begin by describing the task-specific layers: for each task, the hidden representations are first fed to a fully connected layer with rectified linear units (ReLU), whose outputs are passed to another linear transformation to get logits $z = Wh$. During multi-task training, the parameters from both the bottom student network and upper task-specific layers are jointly updated.

Considering one text classification problem, denoted as task $t$, a softmax layer will perform the following operations on the $i^{th}$ dimension of $z$ to get the predicted probability for the $i^{th}$ class:
\begin{equation}
      \text{softmax}(z_{i}^{t}) = \frac{\exp\{z_{i}^{t}\}}{\sum_{j}\exp\{z_{j}^{t}\}}
\end{equation}

According to \citet{ba2014deep}, training the student network on logits will make learning easier. There might be information loss from transferring logits into probability space, so it follows that the teacher model's logits provides more information about the internal model behaviour than its predicted one-hot labels. Then, our distillation objective is to minimize the mean-squared error (MSE) between the student network logits $z_{S}^{t}$ and the teacher's logits $z_{T}^{t}$:
\begin{equation}
    L_{distill}^{t} =  \lVert z_{T}^{t} - z_{S}^{t} \rVert^{2}_{2}
\end{equation}

The training samples are selected from each dataset and packed into task-specific batches. For task $t$, we denote the current selected batch as $b_{t}$. For each epoch, the model running through all the batches equals to attending over all the tasks:
\begin{equation}
    L_{distill} = L_{distill}^{1}+  L_{distill}^{2} + ... + L_{distill}^{t} 
\end{equation}

During training, the teacher model first uses the pretrained BERT model \cite{devlin2019bert} to initialize its parameters of shared layers. It then follows the multi-task refining procedure described in Section \ref{sec:3.1} to update both the bottom shared-layers and upper task-specific layers. 

For student model, the shared parameters are randomly initialized. During training, for each batch, the teacher model first predicts teacher logits. The student model then updates both the bottom shared layer and the upper task-specific layers according to the teacher logits. The complete procedure is summarized in Algorithm \ref{alg: distillalg}.

\begin{algorithm}
    \caption{Multi-task Distillation}
    \label{alg: distillalg}
    \begin{algorithmic}
        \Statex Initialize the shared layers with BERT$_{Large}$ then multi-task refine the teacher model
        \Statex Randomly initialize the student model parameters
        \Statex Set the max number of epoch: $epoch_{max}$
        \Statex \textit{ // Pack the data for $T$ Tasks into batches}
        \For{$t \gets 1$ to $T$}     
            \State{1. Generate augmented data: t$_\text{aug}$}
            \State {2. Pack the dataset $t$ and t$_\text{aug}$ into batch $D_{t}$}
        \EndFor
        \Statex \textit{ // Train the student model}
        \For{$epoch \gets 1$ to $epoch_{max}$}
            \State{1. Merge all datasets:}
            \State{\quad $D = D_{1} \cup D_{2}$ ... $\cup D_{T}$}
            \State{2. Shuffle D}
            \For{$b_{t}$ in D}
                \State{3. Predict logits $z^{T}$ from teacher model}
                \State{4. Predict logits $z^{S}$ from student model}
                \State{5. Compute loss $L_{distill}(\theta)$}
                \State{6. Update student model: }
                \State{\quad $\theta = \theta - \alpha \nabla_{\theta} L_{distill}$}
            \EndFor
        \EndFor

    \end{algorithmic}
\end{algorithm}

\section{An Intuitive Explanation}
In this section we give an intuitive explanation on why using some shared structure during the multi-task training could possibly help. Suppose the samples of the task $T$ are independent and identically distributed $x^{T}, y^{T}\sim \mathcal{P}^{T}_{\mathcal{X}\mathcal{Y}}$, where $x^T$, $y^T$ are the feature and labels of the samples in task $T$ respectively. The joint density can be decomposed as $ p^T(x,y) = p^T(x) p^T(y|x)$. During the discriminative learning process, one tries to estimate the conditional distribution $p^T(\cdot|x)$. For different tasks, $p^T(\cdot|X)$ could be very different. Indeed if there is no connections in $p^T(\cdot|X)$ for different tasks, then it is hard to believe training on one task may help another. However if we assume some smoothness over $p^T(\cdot|X)$, then some connections can be built across tasks.

Without loss of generality, we investigate the case of two tasks. For task $T_1$ and $T_2$, let's assume there exist some common domain of representations $\mathcal{H}$, and two functions: 
$h^{T_1}(x), h^{T_2}(x):\mathcal{X}\mapsto \mathcal{H}$, such that
\begin{align}
p^{T_1}(\cdot|x) &= g^{T_1}\circ h^{T_1}(x), \\
p^{T_2}(\cdot|x) &= g^{T_2}\circ h^{T_2}(x),\\
\forall x_1, x_2,~~&\|h^{T_1}(x_1) - h^{T_2}(x_2)\|\leq \eta \|x_1 - x_2\|, \label{eqn:lipschitz}
\end{align}
where $g^{T}:\mathcal{H}\mapsto \mathcal{Y}^T$ is a function that maps from the common domain $\mathcal{H}$ to the task labels $\mathcal{Y}^T$ for task $T$, $\circ$ denotes function composition, and $\eta$ is a smoothness constant.

The Lipschitz-ish inequality (\ref{eqn:lipschitz}) suggests the hidden representation $h^{T_1}$ on task $T_1$ may help the estimation of $h^{T_2}$, since $h^{T_2} (x_2)$ will be close to $h^{T_1}(x_1)$ if $x_1$ and $x_2$ are close enough. This is implicitly captured if we use one common network to model both $h^{T_1}$ and $h^{T_2}$ since the neural network with ReLU activation is Lipschitz.

\section{Experimental Setup}
\subsection{Datasets}
We conduct the experiments on seven most widely used datasets in the General Language Understanding Evaluation (GLUE) benchmark \cite{wang2018glue}: one sentiment dataset SST-2 \cite{socher2013recursive}, two paraphrase identification datasets QQP \footnote{https://www.quora.com/q/quoradata/First-Quora-Dataset-Release-Question-Pairs} and MRPC \cite{dolan2005automatically}, one text similarity dataset STS-B \cite{cer2017semeval}, and three natural language inference datasets MNLI \cite{williams2018broad}, QNLI \cite{rajpurkar2016squad} and RTE. For the QNLI dataset, version 1 expired on January 30, 2019; the result is evaluated on QNLI version 2.

\subsection{Implementation Details}
We use the released MT-DNN model\footnote{https://github.com/namisan/mt-dnn} to initialize  our teacher model. We further refine the model against the multi-task learning objective for 1 epoch with learning rate set to 5e-4.  The performance of our refined MT-DNN is lower than reported results in ~\citet{liu2019multi}.

The LSTM based student model (\textit{MKD-LSTM}) is initialized randomly. For multi-task distillation, We use the Adam optimizer \cite{kingma2014adam} with learning rates of 5e-4. The batch size is set to 128, and the maximum epoch is 16. We clip the gradient norm within 1 to avoid gradient exploding. The number of BiLSTM hidden units in student model are all set to 256. The output feature size of task-specific linear layers is 512. The Transformer-based student model (\textit{MKD-Transformer}) consists of three layers of Transformers. Following the settings of BERT-PKD, it is initialized with the first three layers parameters from pre-trained BERT-base.

We also fine-tune the multi-task distilled student model for each task. During fine-tuning, the parameters of both shared layers and upper task-specific layers are updated. The learning rate is chosen from $\{1, 1.5, 5\} \times 10^{-5}$ according to the validation set loss on each task. Other parameters remain the same as above. For both teacher and student models, we use WordPiece embeddings \cite{wu2016google} with a 30522 token vocabulary.

\parheaderwithdot{Data augmentation} 
The training data for typical natural language understanding tasks is usually very limited. Larger amounts of data are desirable for the teacher model to fully express its knowledge.
\citet{tang2019distilling} proposes two methods for text data augmentation: masking and POS-guided word replacement.  
We employ the only first masking technique which randomly replaces a word in the sentence with \texttt{[MASK]}, because, as shown in both \citet{tang2019distilling} and our own experiments, POS-guided word replacement does \textit{not} lead to consistent improvements in quality across most of the tasks. Following their strategies, for each word in a sentence, we perform masking with probability $p_{mask} = 0.1$. 
We use the combination of original corpus and augmentation data in distillation procedure. For smaller datasets STS-B, MRPC and RTE, the size of the augmented dataset is 40 times the sizes of the original corpus; 10 times for other larger datasets.

\subsection{Methods and Baselines}

\begin{table*}[t]
\centering
\scalebox{0.92}{
\setlength{\tabcolsep}{6pt}
\begin{tabular}{@{}llcccccccc@{}}
\toprule[1pt]
 \multirow{1}{*}{Model} &
 \multirow{1}{*}{\shortstack[l]{Size \\ \\ \\ \# Param}}
& \multicolumn{1}{c}{SST-2}
& \multicolumn{1}{c}{MRPC}
& \multicolumn{1}{c}{STS-B} 
& \multicolumn{1}{c}{QQP}
& \multicolumn{1}{c}{MNLI-m/mm}
& \multicolumn{1}{c}{QNLI}
& \multicolumn{1}{c}{RTE}
\\ \cmidrule(l){3-9} 
 &&Acc & F$_\text{1}$/Acc & $r$/$\rho$ & F$_\text{1}$/Acc & Acc &Acc&Acc \\ 
 \midrule
MTL-BERT (Teacher) & 303.9M & 94.7 & 84.7/79.7 & 84.0/83.3 & 72.3/89.6& 85.9/85.7 & 90.5 & 77.7 \\
 OpenAI GPT  &116.5M& 91.3 &82.3/75.7&82.0/80.0& 70.3/88.5 & 82.1/81.4 & - & 56.0\\ 
 ELMo  &93.6M&90.4& 84.4/78.0 & 74.2/72.3 & 63.1/84.3 & 74.1/74.5 & 79.8 & 58.9  \\

\midrule
 Distilled BiLSTM  & 1.59M& 91.6 & 82.7/75.6 & 79.6/78.2 & 68.5/88.4 &  72.5/72.4 & - & - \\
 BERT-PKD  & 21.3M & 87.5 & 80.7/72.5 & - & 68.1/87.8 & 76.7/76.3 & 84.7 &58.2 \\
 TinyBERT & 5.0M &92.6 & 86.4/81.2 & 81.2/79.9 & 71.3/89.2 & 82.5/81.8 & 87.7 & 62.9 \\
 BERT$_{\text{EXTREME}}$ & 19.2M & 88.4 & 84.9/78.5& - & - & 78.2/77.7 & - \\

\midrule
\textbf{MKD-LSTM} & 10.2M &91.0 & 85.4/79.7 & 80.9/80.9 & 70.7/88.6 & 78.6/78.4 & 85.4 & \textbf{67.3} \\ 
\textbf{MKD-Transformer} & 21.3M& 90.1 & 86.2/79.8 & \textbf{81.5/81.5} & 71.1/\textbf{89.4} & 79.2/78.5 & 83.5 & 67.0 \\

\bottomrule[1pt]

\end{tabular}}
\caption{Results from the GLUE test server. The first group contains large-scale pretrained language models. The second group lists previous knowledge distillation methods for BERT. Our MKD results based on LSTM and Transformer student model architectures are listed in the last group. The number of parameters doesn't include embedding layer.}
\label{table:main_result}
\end{table*}

Results on test data reported by the official GLUE evaluation server are summarized in Table \ref{table:main_result}. Each entry in the table is briefly introduced below:

\parheaderwithdot{MTL-BERT} We use the multi-task refined BERT (described in Section \ref{sec:3.1}) as our teacher model. We tried to replicate the results of the released MT-DNN \cite{liu2019multi} model.  

\parheaderwithdot{OpenAI GPT} A generative pre-trained Transformer-based language model \cite{radford2018improving}. In contrast to BERT, GPT is auto-regressive, only trained to encode uni-directional context.

\parheaderwithdot{ELMo} \citet{peters2018deep} learns word representations from the concatenation of independently trained left-to-right and right-to-left LSTMs. We report the results of a BiLSTM-based model with bi-attention baseline \cite{wang2018glue} trained on top of ELMo.

\parheaderwithdot{Distilled BiLSTM} \citet{tang2019distilling} distill BERT into a simple BiLSTM. They use different models for single and pair sentences tasks.

\parheaderwithdot{BERT-PKD} The Patient-KD-Skip approach \cite{sun2019patient} which student model patiently learns from multiple intermediate layers of the teacher model. We use their student model consisting of three layers of Transformers.

\parheader{TinyBERT} \citet{jiao2019tinybert} propose a knowledge distillation method specially designed for transformer-based models. It requires a general distillation step which is performed on a large-scale English Wikipedia (2,500 M words) corpus.

\parheaderwithdot{BERT$_{\text{EXTREME}}$} \citet{zhao2019extreme} aims to train a student model with smaller vocabulary and lower hidden state dimensions. Similar to BERT-PKD, they use the same training corpus to train BERT to perform KD.

\begin{table*}[t]
\centering
\scalebox{0.92}{
\setlength{\tabcolsep}{3pt}
\begin{tabular}{@{}llcccccccc@{}}
\toprule[1pt]
\multirow{1}{*}{\#} &
\multirow{1}{*}{Model}
& \multicolumn{1}{c}{SST-2}
& \multicolumn{1}{c}{MRPC}
& \multicolumn{1}{c}{STS-B} 
& \multicolumn{1}{c}{QQP}
& \multicolumn{1}{c}{MNLI-m/mm}
& \multicolumn{1}{c}{QNLI}
& \multicolumn{1}{c}{RTE}
\\ 
\midrule
1 & Biatt LSTM & 85.8 & 80.4/69.9 & 12.24/11.33 & 81.1/86.5 & 73.0/73.7 & 80.3 & 53.1  \\
2 & Single Task Distilled Biatt LSTM & 89.2 & 82.5/72.1 & 20.2/20.0 & 84.6/88.4 & 74.7/75.0 & 82.0 & 52.0 \\
3 & BiLSTM$_{\text{MTL}}$ & 87.5& 83.2/72.8&71.6/72.6&81.6/87.0 & 70.2/71.3 & 75.4 & 56.3 \\
4 & MKD-LSTM  \textit{\scriptsize{Word-level Tokenizer}} & 87.3 & 84.2/75.7 & 72.2/72.6 & 71.1/79.3 & 69.4/70.9 & 75.1 & 54.9 \\
\midrule
5 & \textbf{MKD-LSTM}  & 89.3 & 86.8/81.1 & 84.5/84.5 & 85.2/89.0 & 78.4/79.2 & 83.0 & 67.9 \\

\bottomrule[1pt]

\end{tabular}}
\caption{Ablation studies on GLUE dev set of different training procedures. All models are not fine-tuned. Line 1 is our bi-attentive LSTM student model trained without distillation.  Line 2 is our bi-attentive LSTM student distilled from single task.  Line 3 is the Multi-task distilled BiLSTM. Line 4 is the Multi-task distilled model using word-level tokenizer.}
\label{tab:ablation_1}
\end{table*}

\begin{table*}[t]
\centering
\scalebox{0.92}{
\setlength{\tabcolsep}{4pt}
\begin{tabular}{@{}lcccccccc@{}}
\toprule[1pt]
\multirow{1}{*}{Model}
& \multicolumn{1}{c}{SST-2}
& \multicolumn{1}{c}{MRPC}
& \multicolumn{1}{c}{STS-B} 
& \multicolumn{1}{c}{QQP}
& \multicolumn{1}{c}{MNLI-m/mm}
& \multicolumn{1}{c}{QNLI}
& \multicolumn{1}{c}{RTE}
\\ 
\midrule
 \textbf{Sentiment Task} & \checkmark & & & & &  &  \\
 MKD-LSTM & 89.9 & 81.4/70.8 & 51.2/49.9 &84.9/88.3& 74.3/74.7 &83.2&50.9 \\
 \textbf{PI Tasks} & & \checkmark & &\checkmark & &  &  \\
 MKD-LSTM  & 89.3 & 85.2/77.2 & 83.4/83.3 & 84.9/88.7 & 73.2/73.9 & 83.8 & 59.6 \\
 \textbf{NLI Tasks} & & & & & \checkmark & \checkmark & \checkmark \\
 MKD-LSTM & 90.4 & 87.9/82.1 & 84.1/84.1 & 84.8/88.4 & 77.1/78.1 & 84.5 & 66.8 \\
 \textbf{All Tasks} & \checkmark & \checkmark &\checkmark &\checkmark &\checkmark & \checkmark & \checkmark \\
 MKD-LSTM & 90.5 & 86.9/80.2 & 85.0/84.8  &84.8/89.0& 77.4/78.3 & 84.9 & 68.2 \\

\bottomrule[1pt]

\end{tabular}}
\caption{Ablation experiments on the dev set use different training tasks in multi-task distillation. The results are reported with the original corpus, without augmentation data. The model is fine-tuned on each individual task.}
\label{tab:ablation_2}
\end{table*}

\section{Result and Discussions}
The results of our model are listed as MKD-LSTM and MKD-Transformer in the tables.
\subsection{Model Quality Analysis}
\parheaderwithdot{Comparison with GPT / ELMo} Our model has better or comparable performance compared with ELMo and OpenAI GPT. MKD-LSTM has higher performance than ELMo over all seven datasets: notably 8.4 points for RTE, 8.6 points in Spearman's $\rho$ for STS-B, 7.6 points in F-1 measure for QQP, and 0.6 to 5.6 points higher for other datasets. Compared with OpenAI GPT, MKD-LSTM is 11.3 points higher for RTE and 4 points higher for MRPC. 

\parheaderwithdot{Comparison with Distilled BiLSTM / BERT-PKD} While using the same Transformer layers and same amount of parameters, MKD-Tranformer significantly outperforms BERT-PKD by a range of $0.4\sim9.1$ points. MKD-LSTM leads to significant performance gains than BERT-PKD while using far less parameters, and compensate for the effectiveness loss of Distlled BiLSTM.

\parheaderwithdot{Comparison with TinyBERT / BERT$_{\text{EXTREME}}$} These two approaches both use the large-scale unsupervised text corpus, same as the ones to train the teacher model, to execute their distillation process. However, we only use the data within downstream tasks. There are two caveats for their methods: (1) Due to massive training data, KD still requires intensive computing resources, e.g. BERT$_{\text{EXTREME}}$ takes 4 days on 32 TPU cores to train their student model. 
(2) The text corpus to train the teacher model is not always available due to data privacy. Under some conditions we can only access to the pretrained models and their approach are not applicable. 

While not resorting to external training data, our model has the best performance across the state-of-the-art KD baselines (i.e., BERT-PKD). It also achieves comparable performance compared to intensively trained KD methods (i.e, BERT$_{\text{EXTREME}}$) on external large corpus.

\subsection{Ablation Study}
We conduct ablation studies to investigate the contributions of: (1) the different training procedures (in Table \ref{tab:ablation_1}); (2) Different training tasks in multi-task distillation (in Table \ref{tab:ablation_2}). We also compare the inference speed of our models and previous distillation approach. (in Table \ref{tab: speed analysis}). The ablation studies are all conducted on LSTM-based student model since it has the advantage of model size and inference speed compared to Transformers.

\parheader{Do we need attention in the student model?}
Yes. \citet{tang2019distilling} distill BERT into a simple BiLSTM network. Results in Table \ref{table:main_result} demonstrates that our model is better than Distilled BiLSTM and achieves an improvement range of $2.2 \sim 6.1$ points across six datasets. To make fair comparison, we also list the results of multi-task distilled BiLSTM in Line 3 in Table \ref{tab:ablation_1}. It's obvious that Line 5, which is the model with bi-attentive mechanism, significantly outperform Line 3. We surmise that the attention module is an integral part of the student model for sequence modeling.

\parheader{Better vocabulary choices?}
WordPiece works better than the word-level tokenizers in our experiments. The WordPiece-tokenized vocabulary size is 30522, while the word-level tokenized vocabulary size is much larger, along with more unknown tokens. WordPiece effectively reduces the vocabulary size and improves rare-word handling. The comparison between Line 4 and Line 5 in Table \ref{tab:ablation_1} demonstrates that the method of tokenization influences all the tasks.

\parheader{The influence of MTL in KD?} 
The single-task distilled results are represented in Line 2 of Table \ref{tab:ablation_1}. Compared with Line 5, all the tasks benefit from information sharing through multi-task distillation. Especially for STS-B, the only regression task, greatly benefit from the joint learning from other classification tasks. 

We also illustrate the influence of different number of tasks for training. In Table \ref{tab:ablation_2}, the training set incorporates tasks of the same type individually. Even for the tasks which are in the training sets, they still perform better in the all tasks training setting. For example, for RTE, the \textit{All Tasks} setting increases 1.4 points than \textit{NLI Tasks} setting. For other training settings which RTE is excluded from training set, \textit{All Tasks} leads to better performance.

\begin{table}[t]
\centering
\scalebox{0.7}{
\setlength{\tabcolsep}{4.0pt}
\begin{tabular}{@{}lccccccc@{}}
\toprule[1pt]
& \multicolumn{1}{c}{Distilled BiLSTM}
& \multicolumn{1}{c}{BERT-PKD} 
& \multicolumn{1}{c}{TinyBERT}
& \multicolumn{1}{c}{MKD-LSTM}
\\ 
\midrule
Inf. Time & 1.36 & 8.41 & 3.68 & 2.93   \\
\bottomrule[1pt]
\end{tabular}}
\caption{The inference time (in seconds) for baselines and our model. The inference is performed on QNLI training set and on a single NVIDIA V100 GPU.}  
\label{tab: speed analysis}
\end{table}

\subsection{Inference Efficiency}
To test the model efficiency, we ran the experiments on QNLI training set. We perform the inference on a single NVIDIA V100 GPU with batch size of 128, maximum sequence length of 128. The reported inference time is the total running time of 100 batches. 

From Table \ref{tab: speed analysis}, the inference time for our model is 2.93s. We re-implemented Distilled BiLSTM from \citet{tang2019distilling} and their inference time is 1.36s. For fair comparison, we also ran inference procedure using the released BERT-PKD and TinyBERT model on the same machine.
Our model significantly outperforms Distilled BiLSTM with same magnitude speed. It also achieves comparable results but is faster in efficiency compared with other distillation models.

\section{Conclusion}
In this paper, we propose a general framework for multi-task knowledge distillation. The student is jointly distilled across different tasks from a multi-task refined BERT model (teacher model). We evaluate our approach on Transformer-based and LSTM-based student model. Compared with previous KD methods using only data within tasks, our approach achieves better performance. In contrast to other KD methods using large-scale external text corpus, our approach balances the problem of computational resources, inference speed, performance gains and availability of training data.

\bibliography{emnlp2020}
\bibliographystyle{acl_natbib}

\end{document}